\documentclass{sig-alternate}
\usepackage[square,comma,sort&compress,numbers]{natbib}
\begin{document}
%
\conferenceinfo{PCSC 2016}{March 16-18, 2016, Puerto Princesa City, Palawan.}
\CopyrightYear{2016} 
\crdata{x-xxxxx-xx}  

\title{Improved Sampling Techniques for\\Learning an Imbalanced Data Set}

\numberofauthors{1}
\author{
\alignauthor
Maureen Lyndel C. Lauron and Jaderick P. Pabico\\
       \affaddr{Institute of Computer Science}\\
       \affaddr{University of the Philippines Los Ba\~nos}\\
       \affaddr{College 4031, Laguna}
}

\maketitle
\begin{abstract}
This paper presents the performance of a classifier built using the stackingC algorithm in nine different data sets. Each data set is generated using a sampling technique applied on the original imbalanced data set. Five new sampling techniques are proposed in this paper (i.e., SMOTERandRep, Lax Random Oversampling, Lax Random Undersampling, Combined-Lax Random Oversampling Undersampling, and Combined-Lax Random Undersampling Oversampling) that were based on the three sampling techniques (i.e., Random Undersampling, Random Oversampling, and Synthetic Minority Oversampling Technique) usually used as solutions in imbalance learning. The metrics used to evaluate the classifier's performance were F-measure and G-mean. F-measure determines the performance of the classifier for every class, while G-mean measures the overall performance of the classifier. The results using F-measure showed that for the data without a sampling technique, the classifier's performance is good only for the majority class. It also showed that among the eight sampling techniques, RU and LRU have the worst performance while other techniques (i.e., RO, C-LRUO and C-LROU) performed well only on some classes. The best performing techniques in all data sets were SMOTE, SMOTERandRep, and LRO having the lowest F-measure values between 0.5 and 0.65.  The results using G-mean showed that the oversampling technique that attained the highest G-mean value is LRO (0.86), next is C-LROU (0.85), then SMOTE (0.84) and finally is SMOTERandRep (0.83). Combining the result of the two metrics (F-measure and G-mean), only the three sampling techniques are considered as good performing (i.e., LRO, SMOTE, and SMOTERandRep)
\end{abstract}

\section{Introduction}
In data mining, to have an extreme imbalanced data set is inevitable such as the case of the classification problem in this study. It is natural for the data set to have a skewed distribution because there are far more continuing students (COS) than students that go on leave of absence (LOA). The class with the most number of instances (i.e., COS) is called the majority class while the class with the small number of instances (i.e., LOA) is called the minority class. In the case this study, there are twelve minority classes (refer to Figure 1) thus the sampling techniques were used as solutions to the multi-class imbalance learning problem. 
	
The problem with an extreme imbalance data set is the skewed distribution of the data that makes the learning algorithms ineffective, especially in predicting minority classes. This problem has drawn enough attention from both industry and academia, thus many state-of-the-art solutions now exists such as sampling techniques, cost sensitive methods, and kernel-based methods~\citep{he-2009}. Even if some empirical studies have shown that cost sensitive and kernel-based methods are superior than sampling techniques, sampling techniques still also dominates the current researches for imbalanced learning because of its flexibility for any learning algorithm. Meanwhile, the two techniques can be used only with a specific learning algorithm, for example, kernel-based methods is implemented with support vector machines and some cost sensitive methods with adaptive boosting~\citep{he-2009}.
	
There are many machine learning algorithms developed already for the classification task but each algorithm typically suits some classification problems better than others; and this is why there is no universal data mining method for all problem types~\citep{fayyad-1996}. With the claim that all classification algorithms are equally alike with each having its own advantages and disadvantages, a question on whether a new algorithm that mixes some of the advantages of the algorithms performs better than individual algorithms was investigated. This new algorithm composed of different learning algorithms is called ensemble learning and it was verified using multiple data sets that an ensemble learner has a better performance than an individual learning algorithm~\citep{lauron-2016}. The ensemble learning algorithm used in building a classifier in this study is stackingC and the combined learning algorithms were support vector machines (SVM), k-nearest neighbor (k-NN), C4.5, classification and regression trees (CART), and multilayer perceptron (MP). This combination of learning algorithms was used because it was found to be the near optimal combination of learning algorithms for stackingC ensemble classifier~\citep{lauron-2016}.
	
The classification problem of this study concerns the classification of the incoming freshmen student of the University of the Philippines Los Ba\~{n}os (UPLB) if whether they will shift course, transfer to another school or UP campus, go on absence without leave (AWOL), be a continuing student, and so forth using their available data such as their high school (HS) grades, HS type, UP College Admission Test result, family income, address, etc.

\section{Sampling Techniques}
\label{sec:samptech}
\looseness-1Sampling techniques as a solution to the imbalance learning problem simply modifies an imbalanced data set so that it will have a somewhat balanced distribution. Some sampling techniques commonly used in imbalance learning are random undersampling (RU), random oversampling (RO), and synthetic minority oversampling technique (SMOTE)~\citep{he-2009}. 
\subsection{Sampling Techniques in the Literature}
\subsubsection{Random Undersampling}
The mechanics of RU as its name implies, randomly removes instances from all classes except from the class with the least number of instances (most minority class) until the number of instances for all the classes is equal. 
\subsubsection{Random Oversampling}
While undersampling removes data from the original data, RO adds data by randomly replicating the instances of the twelve minority classes until the number of instances for all the classes are equal. 
\subsubsection{Synthetic Minority Oversampling Technique}
The SMOTE just like RO also adds new data but in a different manner~\citep{chawla-2002}. SMOTE creates artificial data based on feature space similarities between existing instances of the minority class. This technique uses the k-nearest neighbor algorithm to find the k-nearest neighbors of an instance $x_{i}$ in the minority class. To create a synthetic instance $x_\mathrm{new}$, a random instance $x^{'}$ is chosen from the k-nearest neighbors, then the corresponding feature vector difference is multiplied with a random number between [0,1] and finally this vector is added to $x_{i}$ 
\begin{equation}
x_\mathrm{new} = x_{i} + (x^{'}- x_{i}) \times \delta
\end{equation}
where $x_{i}$ is a minority instance, $x^{'}$ is the randomly chosen k-nearest neighbor, and $\delta$ is a random number between 0 and 1.

\subsection{Proposed Sampling Techniques}
The five new sampling techniques proposed in this paper are based on the abovementioned sampling techniques. 
\subsubsection{SMOTERandRep}
The first technique is called SMOTERandRep, a short name for SMOTE Random Replacement. This technique is based on the idea of SMOTE because it also oversamples the minority class and create synthetic instances. Given an instance $x_{i}$ of a minority class and its k-nearest neighbors, it creates a new instance by copying the features of $x_{i}$ and randomly replacing $\log_{2}a$ attribute values with the values from the randomly selected k-nearest neighbors where $a$ is the total number of attributes and k value is also equal to $\log_{2}a$. 
\subsubsection{Lax Random Undersampling}
The second technique is called Lax Random Undersampling (LRU). This technique is based on the idea of RU and the difference is instead of aiming for having an equal number of instances for the minority class and majority class, it randomly delete instances from the majority class until the number of instances in this class is equal to the number of instances of the minority class with the most number of instances.
 \subsubsection{Lax Random Oversampling}
 The third technique is called Lax Random Oversampling (LRO) and as its name suggests, is based on the idea of RO. This technique differs with RO because instead of having an equal number of instances for the minority class and majority class, it randomly replicates the minority instances until the minority class with the most number of instances is replicated once. 
 \subsubsection{Combined-LROU}
 The fourth and fifth techniques are simply the combination of the second and third techniques (LRO and LRU) and that is why it is called Combined-LROU (C-LROU) and Combined-LRUO (C-LRUO). The difference of the two techniques is the order of combining LRO and LRU, for example, C-LROU first applies LRO on the original data set, then LRU is applied on the resulting data set. 
 \subsubsection{Combined-LRUO}
 As for C-LRUO, it first applies LRU on the original data set and after that LRO is applied on the resulting data set.

\section{Methodology}
\label{sec:methodology}

The student data used in this study for the UPLB classification problem was gathered from the Office of the University Registrar (OUR) of UPLB and from five different colleges of the university. The collected data identified 13 possible classification of an instance of a student and these are shiftee (SHC), transfer to another UP campus (TUP), transfer to another school (TAS), scholastic delinquency warning (SDW), scholastic delinquency probationary (SDP), scholastic delinquency dismissal (SDD), scholastic delinquency permanent disqualification (SPQ), readmission of disqualified and dismissed students (RAM), absence without leave (AWL), leave of absecence (LOA), honorable dismissal (HDM), scholastic delinquency warning and probationary (SWP), and continuing student (COS). A total of 2297 instances of students were collected and the distribution of the data per class is shown in Figure~1.
\begin{figure*}[t]
\centerline{\includegraphics[width=14cm]{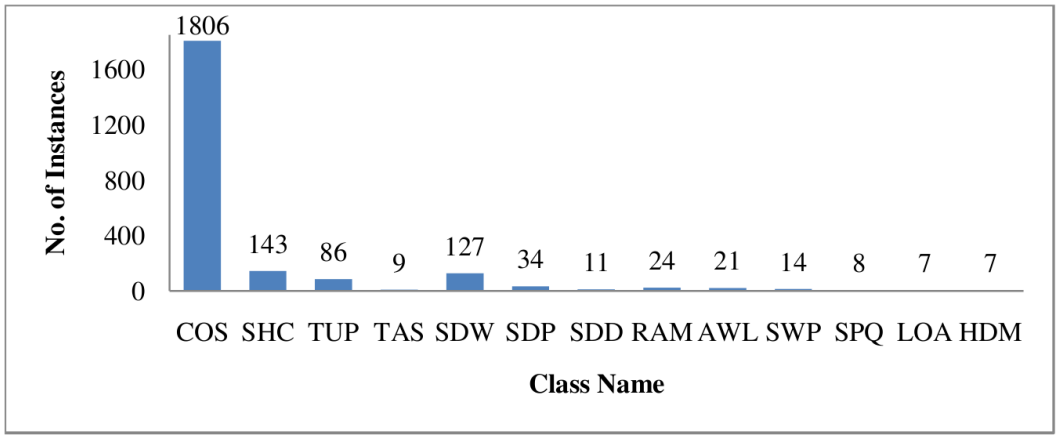}}
\caption{Distribution of the original students' data on 13 classes.}
  \label{fig:origdata}
\end{figure*}

Waikato Environment for Knowledge Analysis (WEKA) is an open source (available under GNU General Public License) and cross-platform (written in Java) data mining software for data mining tasks~\citep{hall-2009}. WEKA collects and automates state-of-the-art data mining algorithms thus, its implementation of stackingC ensemble learning algorithm as well as the implementations of other algorithms (i.e., SVM, MP, CART, C4.5, and k-NN) was used in this study. WEKA's implementation of SMOTE technique was also used in this study. The evaluation metrics used to measure the performance of the classifier were F-measure and G-mean together with stratified tenfold cross-validation as the validation method. Both metrics are widely used metric; the F-measure metric shows the classifier's performance for every class while G-mean measures the overall performance of a classifier~\citep{wang-2012}. The maximum value for both metrics is one. WEKA has also provided the functionality to compute the F-measure value of a generated classifier and it was also used in this study. The G-mean metric, on the other hand, can be easily computed using recall value performance of the classifier for every class (also available in WEKA) and the formula is
\begin{equation}
	\mathrm{G\text{-}mean} = \left(\Pi_{i=1}^{c}rc_{i}\right)^{\frac{1}{c}}
\end{equation}
where $c$ is the total number of classes and  $rc_{i}$is the recall value of the class.

After applying the eight sampling techniques on the original student data, the new data distributions per sampling technique is shown in Figures~2--6. The distribution of the data for both SMOTE and SMOTERandRep is the same as shown in Figure~2.  
\begin{figure*}[t]
\centerline{\includegraphics[width=14cm]{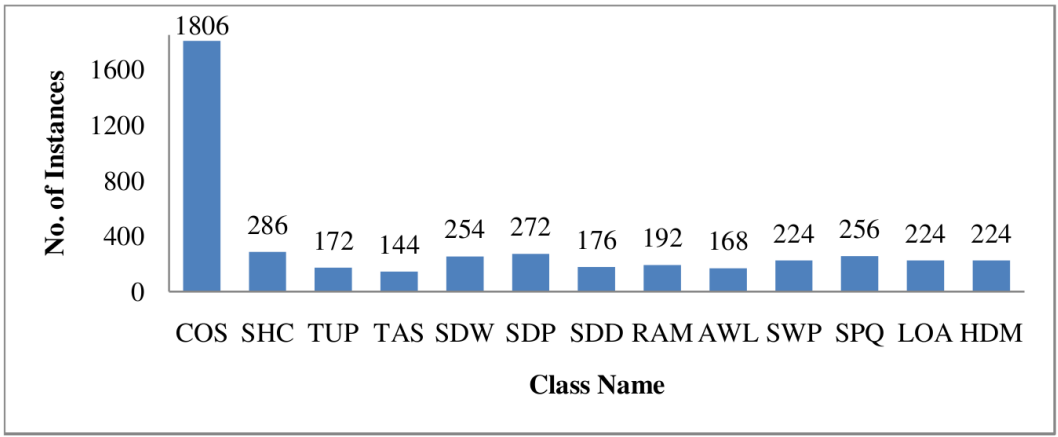}}
\caption{Distribution of the students' data on 13 classes after applying either SMOTE or SMOTERandRep sampling technique.}
  \label{fig:smotesdata}
\end{figure*}

\begin{figure*}[t]
\centerline{\includegraphics[width=14cm]{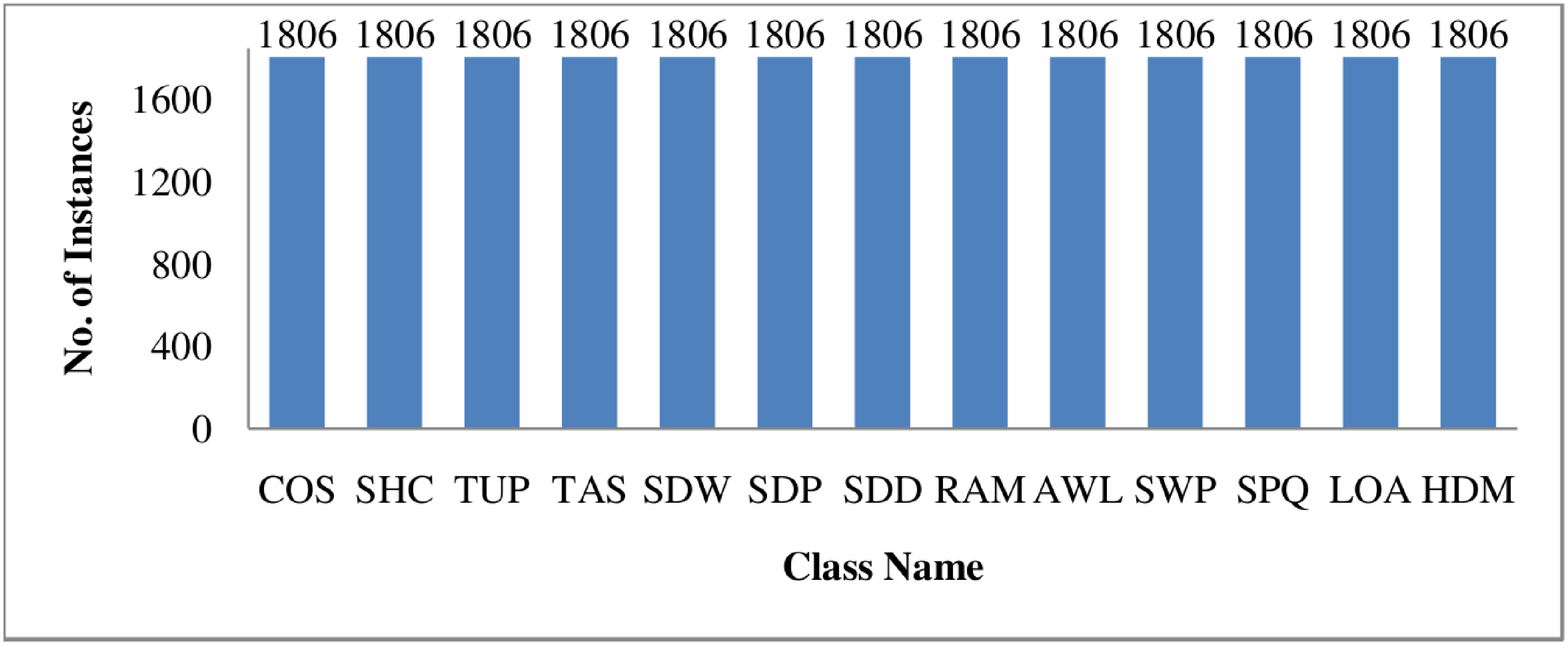}}
\caption{Distribution of the students' data on 13 classes after applying RO.}
  \label{fig:rodata}
\end{figure*}

\begin{figure*}[t]
\centerline{\includegraphics[width=14cm]{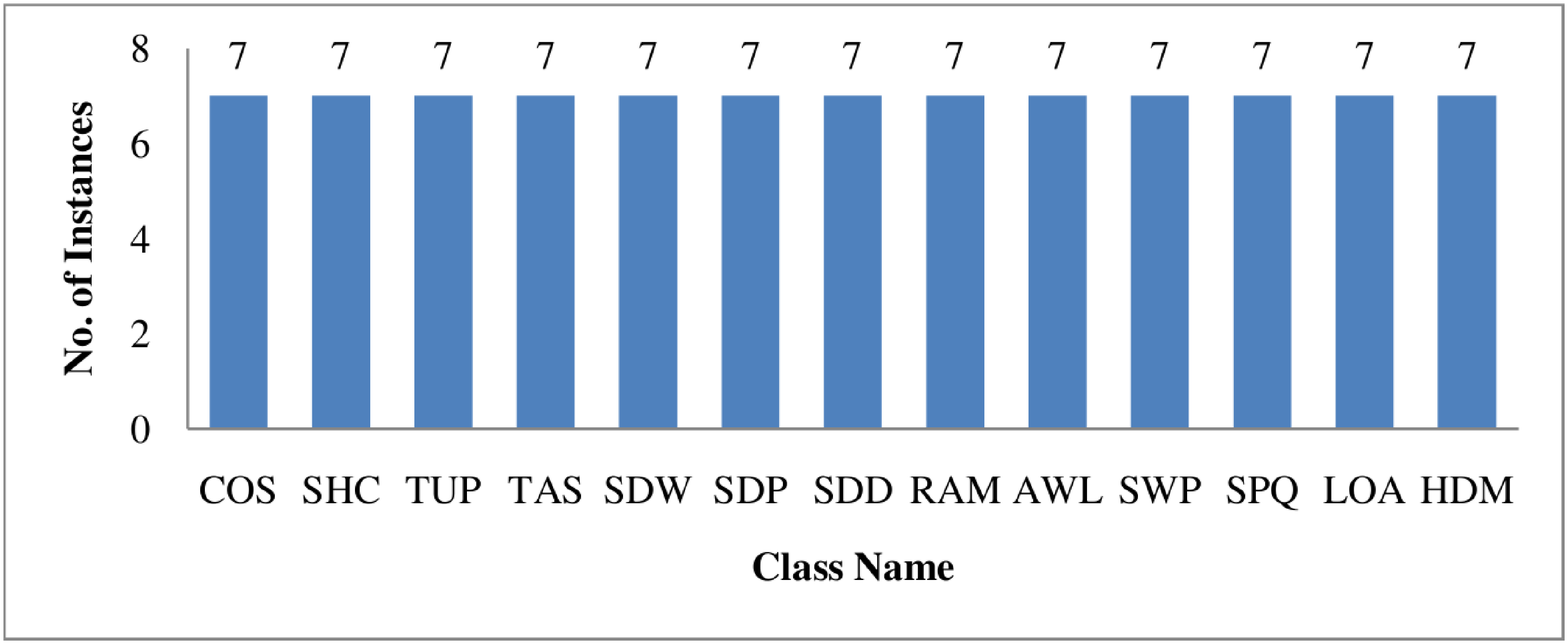}}
\caption{Distribution of the students' data on 13 classes after applying RU.}
  \label{fig:rudata}
\end{figure*}

\begin{figure*}[t]
\centerline{\includegraphics[width=14cm]{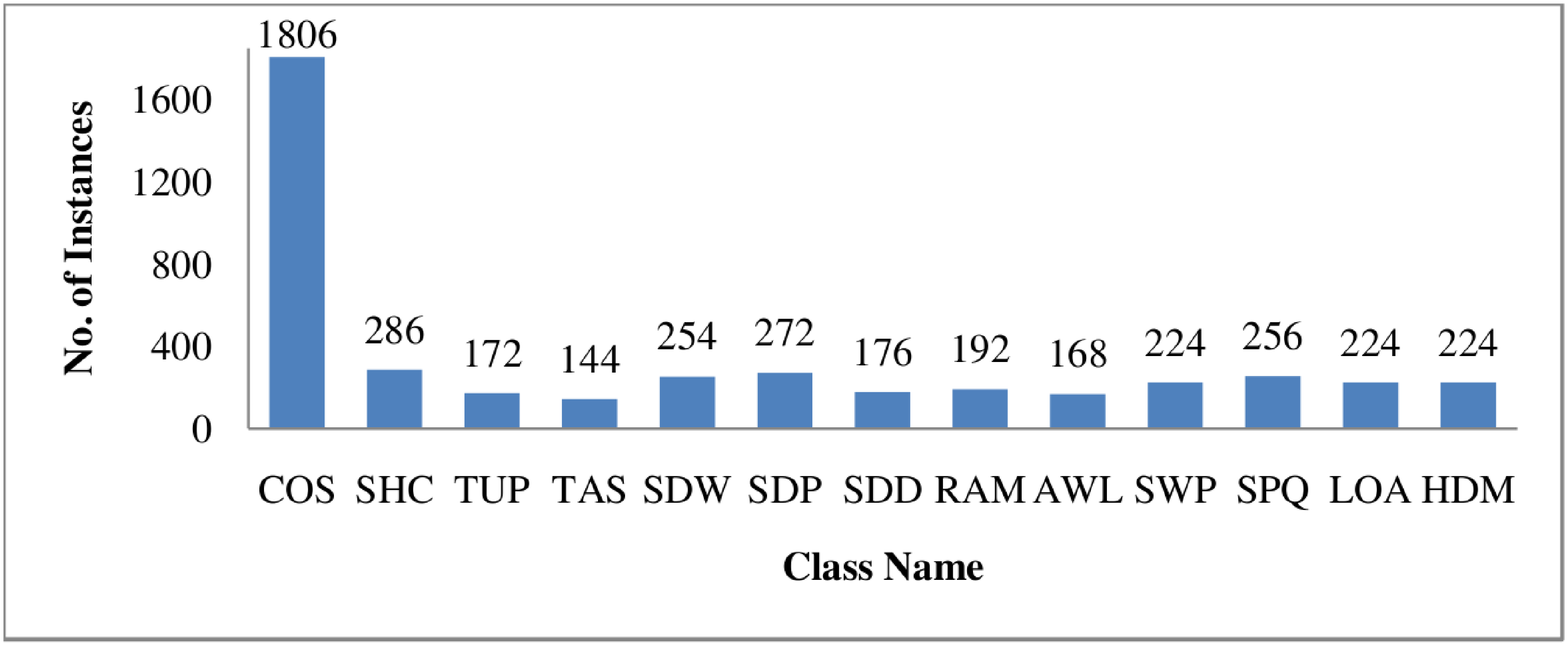}}
\caption{Distribution of the students' data on 13 classes after applying LRO.}
  \label{fig:lrodata}
\end{figure*}

\begin{figure*}[t]
\centerline{\includegraphics[width=14cm]{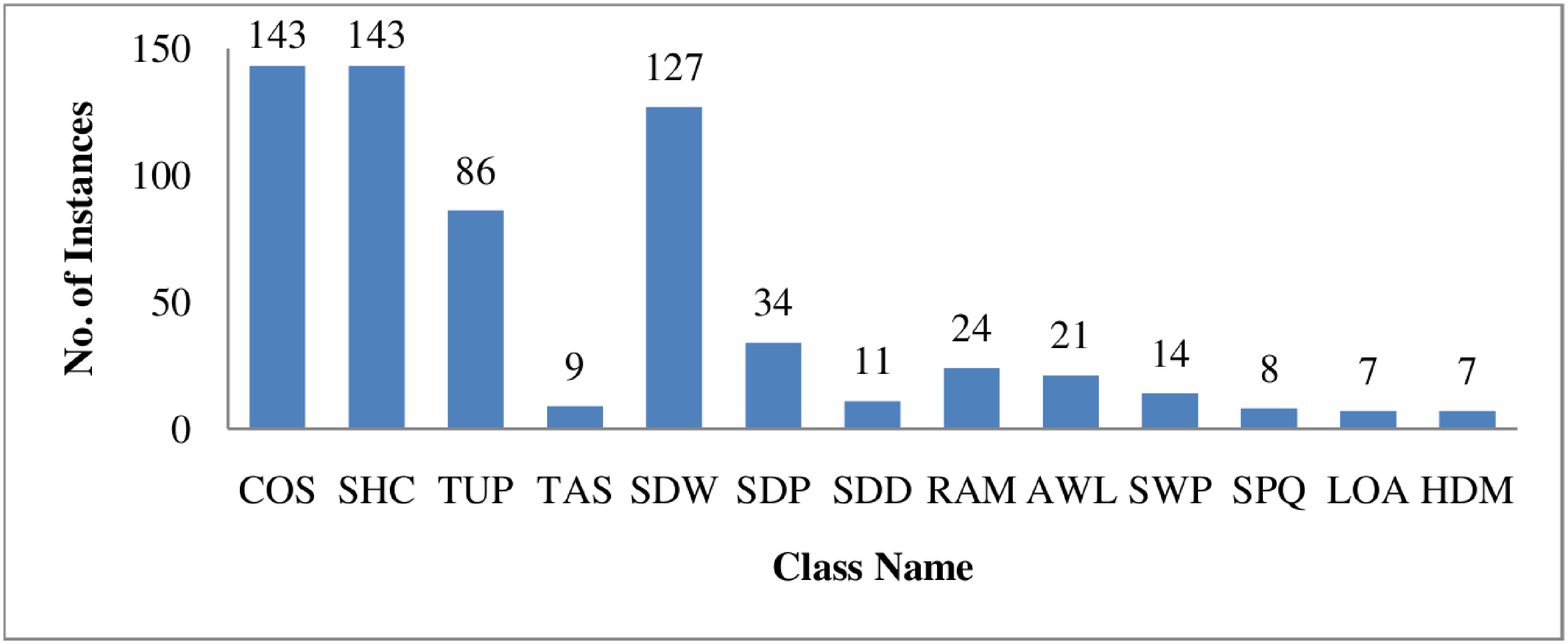}}
\caption{Distribution of the students' data on 13 classes after applying LRU.}
  \label{fig:lrudata}
\end{figure*}

\begin{figure*}[t]
\centerline{\includegraphics[width=14cm]{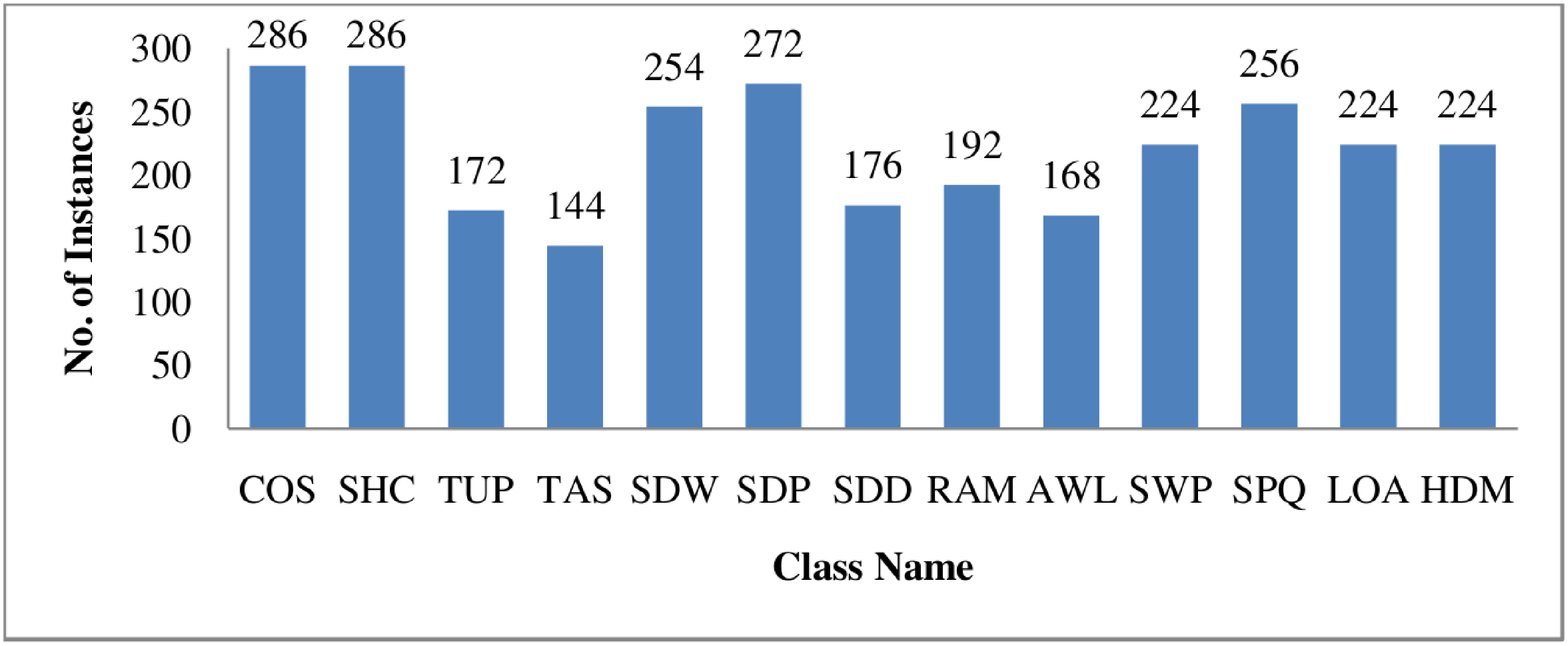}}
\caption{Distribution of the students' data on 13 classes after applying C-LROU.}
  \label{fig:clroudata}
\end{figure*}
\begin{figure*}[t]
\centerline{\includegraphics[width=14cm]{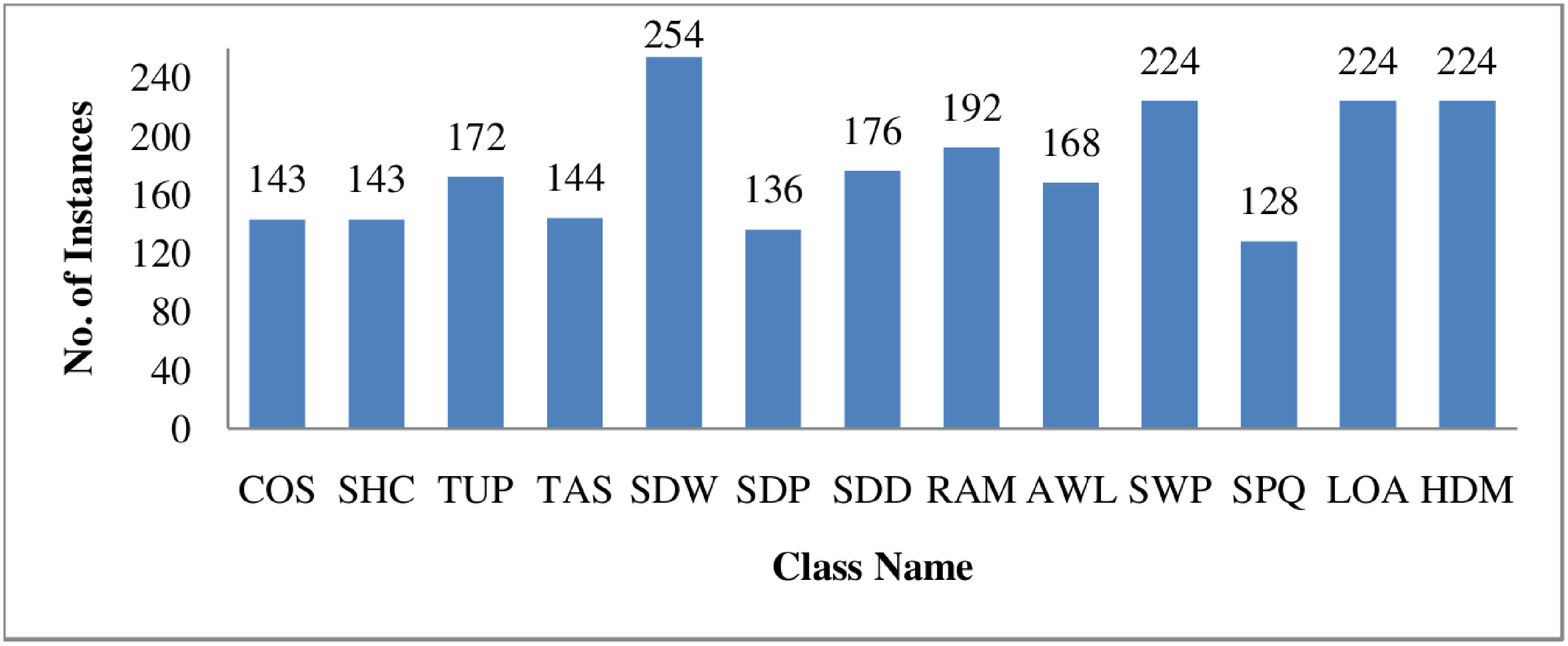}}
\caption{Distribution of the students' data on 13 classes after applying C-LRUO.}
  \label{fig:clruodata}
\end{figure*}

\section{Results}
\label{sec:result}
As expected, the classifier's performance using the original data is good for the majority class (COS) alone as shown in Figure 9. Among the eight sampling techniques, RU and LRU showed the worst performance for all the classes while other techniques (i.e., RO, C-LRUO, and C-LROU) performed well only on some of the classes. The best performing sampling techniques for all the classes were SMOTE, SMOTERandRep, and LRO having the lowest F-measure class values (SDW) of 0.65, 0.52, and 0.61 respectively. The result implies that sampling techniques indeed improves the performance of the classifier and that oversampling techniques are preferable than undersampling techniques.

\begin{figure*}[t]
\centerline{\includegraphics[width=18cm]{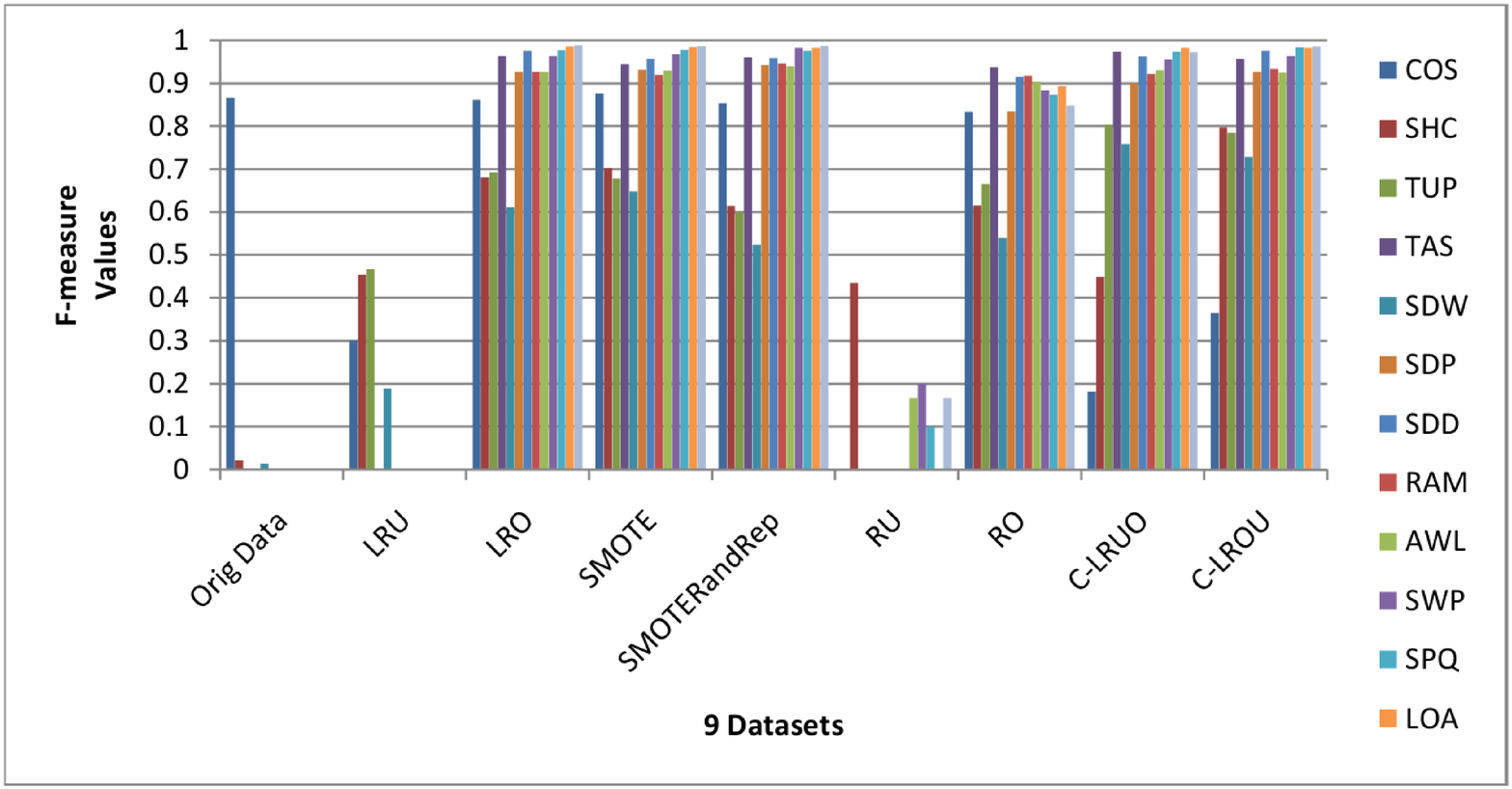}}
\caption{Classifier's performance over nine data sets using F-measure metric.}
  \label{fig:fmeasure}
\end{figure*}

As for the overall performance of the classifier, the lowest G-mean values were of the original data, LRU, and RU as shown in Figure 10. The top sampling techniques with the highest G-mean values were LRO (0.86), C-LROU (0.85), SMOTE (0.84), and SMOTERandRep (0.83). The G-mean results are consistent with the F-measure results except for C-LROU. This still means that the C-LROU technique cannot be considered as a sampling technique with a good performance because even if it has the highest F-measure values on all classes (except the majority class), it has jeopardized the performance of the classifier on the majority class. 
\begin{figure*}[t]
\centerline{\includegraphics[width=14cm]{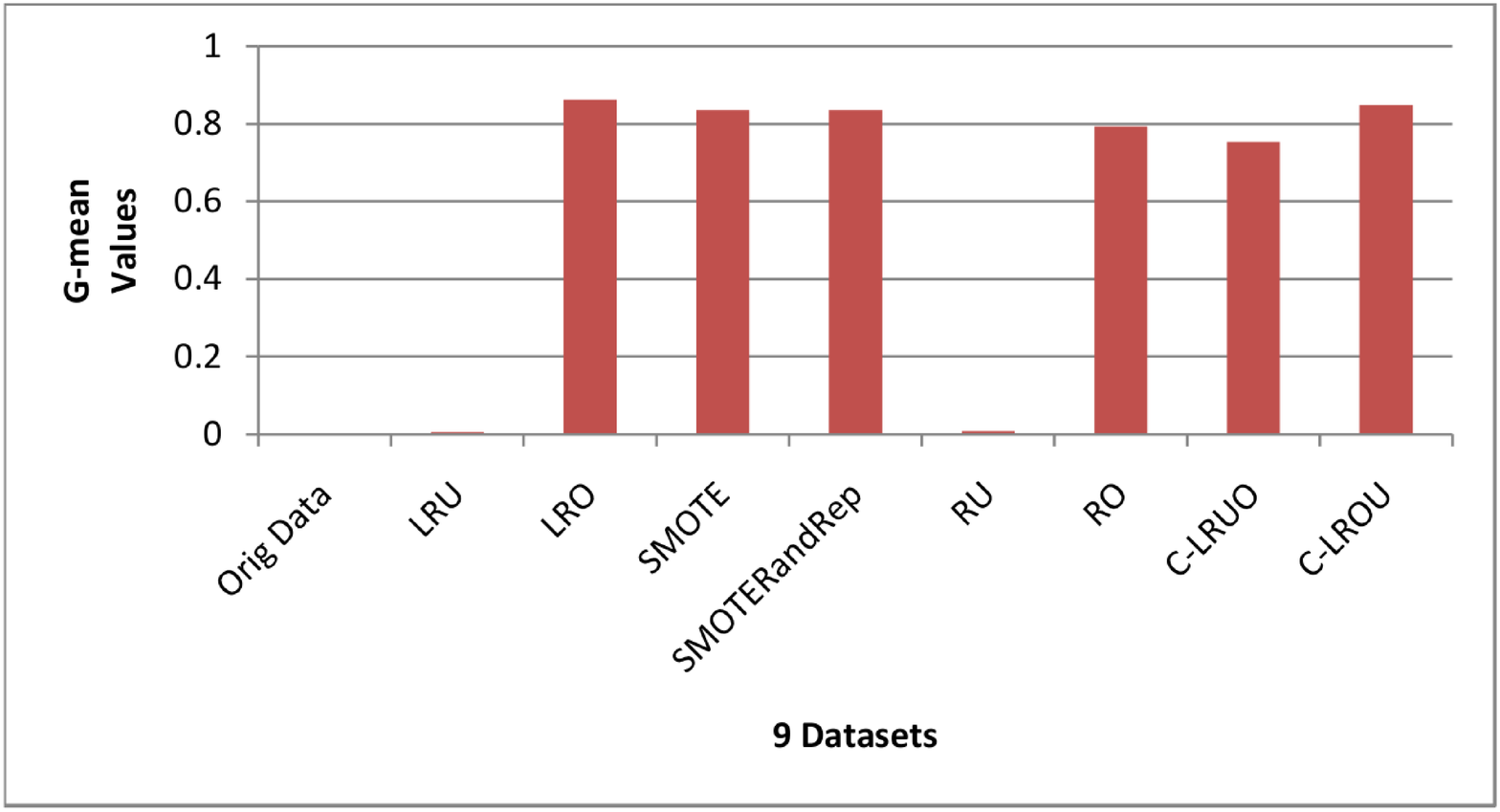}}
\caption{Classifier's performance over nine data sets using G-mean metric.}
  \label{fig:gmean}
\end{figure*}

\section{Conclusion}
\label{sec:conc}
The results showed that indeed sampling techniques improved the performance of the classifier for all minority classes. Between the two techniques (RO and RU), the results also showed that RO is preferable. The reason of this is obvious, since RU removes instances of the majority class, as the classifier is generated, it missed important concepts about the majority class resulting to a better performance on some minority classes but jeopardized performance on the majority class. The result of this study also showed that LRO, SMOTE, SMOTERandRep were the best performing techniques. To further verify this result, it is suggested that the next iterations of this study make use of larger multi-class imbalanced data set and use hold and wait validation method, instead of stratified tenfold cross-validation.
\bibliographystyle{plainnat}
\bibliography{improved-sampling-techniques}

\balancecolumns 
\end{document}